\documentclass[tikz,border=3mm]{article}
\usepackage{graphicx}
\usepackage{tikz}
\usetikzlibrary{shapes.geometric, arrows}
\usepackage{tikzit}
\usepackage{tikzscale}
\tikzstyle{Outline Node}=[fill=white, draw=black, shape=circle]

\tikzstyle{new edge style 0}=[->]

\usepackage{amsfonts}
\usetikzlibrary{matrix}
\usepackage{amsmath}
\usepackage[toc,page]{appendix}
\usepackage{subcaption}
\usepackage[english]{babel}

\usepackage[letterpaper,top=2cm,bottom=2cm,left=3cm,right=3cm,marginparwidth=1.75cm]{geometry}
\usepackage{authblk}

\providecommand{\keywords}[1]
{
  \small	
  \textbf{\textit{Keywords---}} #1
}

\newcommand{\commentbyKG}[1]{}

\def\U{\mathcal{U}}
\def\V{\mathcal{V}}

\numberwithin{equation}{section}

\begin{document}
\title{Cooperative Multi-Agent Reinforcement Learning for Inventory Management}
 \author[1]{Madhav Khirwar}
\author[1]{Karthik S. Gurumoorthy}
\author[1]{Ankit Ajit Jain}
\author[1]{Shantala Manchenahally}

 \affil[1]{Walmart Global Tech, Bangalore, India}
\affil[ ]{{\{madhav.khirwar, karthik.gurumoorthy, ankit.ajit.jain, shantala.manchenahally\}@walmart.com}}

\date{}
\maketitle
\begin{abstract}
With Reinforcement Learning (RL) for inventory management (IM) being a nascent field of research, approaches tend to be limited to simple, linear environments with implementations that are minor modifications of off-the-shelf RL algorithms. Scaling these simplistic environments to a real-world supply chain comes with a few challenges such as: minimizing the computational requirements of the environment, specifying agent configurations that are representative of dynamics at real world stores and warehouses, and specifying a reward framework that encourages desirable behavior across the whole supply chain. In this work, we present a system with a custom GPU-parallelized environment that consists of one warehouse and multiple stores, a novel architecture for agent-environment dynamics incorporating enhanced state and action spaces, and a shared reward specification that seeks to optimize for a large retailer's supply chain needs. Each vertex in the supply chain graph is an independent agent that, based on its own inventory, able to place replenishment orders to the vertex upstream. The warehouse agent, aside from placing orders from the supplier, has the special property of also being able to constrain replenishment to stores downstream, which results in it learning an additional allocation sub-policy. We achieve a system that outperforms standard inventory control policies such as a base-stock policy and other RL-based specifications for 1 product, and lay out a future direction of work for multiple products.\\\\
\end{abstract}

\section{Introduction}
\label{sec:intro}
Inventory management (IM) is the process of overseeing and controlling the flow of goods from the point of acquisition to the point of sale. The goal of IM is to ensure that an organization has the right products, in the right quantities, at the right time, and at the right place to meet customer demand while minimizing operation costs. A retail giant may consist of hundreds of stores dispersed through a vast geographic area, with each offering thousands of products. For each store, the inventory for these products is supplied by the warehouse to which it is mapped to in the supply chain topology (refer Fig.~\ref{topology}). The warehouse in turn receives its replenishment from a dedicated supplier. The replenishment orders are fulfilled by trailers at regular intervals of time. Stores and warehouses are responsible for maintaining enough product inventory to cover for unexpected delays in replenishment for a few time periods.

The process of managing inventory involves various trade-offs, including maintaining inventory levels at the store and minimizing costs associated with holding inventory. Some key concepts in IM are: stock-out, holding costs, and lead time. A stock-out at a store or warehouse occurs if the inventory for a particular product at the store goes to $0$. This is undesirable as it not only leads to lost sales but also poor customer experience. Holding costs are incurred by stores and warehouses to maintain on-hand inventory of products. These are comprised of quantities such as electricity costs for the store, refrigeration costs for food items, and storage area maintenance costs. Since these costs scale with the amount of inventory being kept on-hand, it is sub-optimal for a store or warehouse to keep the maximum possible amount of inventory on-hand. Lead time refers to the time it takes for a supplier or warehouse to deliver a replenishment order to a warehouse or store, respectively. 

In this work, we demonstrate that our proposed system is able to successfully manage inventory for a single product across a simulated supply chain that mimics the complexities of its real-world counterpart. We discuss scaling the products by including results for 10 products, and propose future research directions to address challenges associated with simultaneously managing inventory for thousands of products. For this paper we define an inventory management problem as: \emph{Given the distribution of demand for each product at each store, find optimal replenishment quantities for each product at each store and at the warehouse, such that over a specified number of time periods, system-wide profit is maximized (equivalently, cost is minimized)}.

\subsection{Reinforcement Learning for IM}
Previous literature makes a case for an RL-based approach to building IM solutions over traditional optimization approaches, as RL systems have the ability to consider long-term trajectories of the future, which day-to-day heuristic optimization systems lack~\cite{or-gym,DBLP:journals/corr/abs-2006-04037,peng2019deep,oroojlooyjadid2022deep}. In fact, this enables RL systems to operate without an additional demand forecasting model, as they implicitly learn to predict customer demand during training. We demonstrate this property in our experiments, where the inventory policies learnt by RL agents are superior compared to optimisation methods such as base-stock policy (BSP)~\cite{anbazhagan2013base}, without requiring an additional demand forecasting model.

Multi-agent RL (MARL) is a machine learning framework that uses multiple agents to learn and adapt. MARL may be a better choice for IM in a supply chain than heuristics-based optimization approaches, which often assume static supply chain properties like constant demand or fixed lead times for procurement. In a supply chain, product demand can fluctuate, and supplier delays and transportation issues can affect lead times. MARL algorithms can learn and adapt to environmental changes in real time without these assumptions (when trained on historical data)~\cite{stranieri2022deep}. This makes them more resilient to supply chain uncertainty and variability. Unlike traditional optimization methods, MARL algorithms can be implemented in a decentralized manner with each supply chain agent making decisions based on local information. This improves supply chain flexibility as minor changes to a supply chain's topology do not require the entire system to be re-trained from scratch.
	
\subsection{Contributions}
The main contributions of this paper are: (i) a novel multi-agent architecture for IM where the warehouse agent has enhanced state and action spaces enabling it to effectively learn an allocation policy especially when it does not have sufficient inventory to meet all the store requests, (ii) a novel reward specification to encourage system-wide cooperation where all agents in the supply chain share the same reward that is calculated for each time period based on the dynamics of the supply chain as a whole, and (iii) a CuPy-parallelized environment that can process all products in constant time, subject to GPU constraints. Henceforth we refer to our enhanced warehouse, shared reward, multi-agent RL system as Cooperative MARL (CMARL).
	
\section{Related Work}
\label{sec:related_work}
 RL techniques frame the IM problem as a Markov Decision Process, with the state space being the current levels of inventory and the action space being the replenishment quantity for each item to be ordered at each time step. This approach has been gaining popularity on small-scale proof-of-concept environments in recent years. For instance, \cite{oroojlooyjadid2022deep} extends deep Q-networks (DQN) to solve a decentralized variant of the beer game problem \cite{d2009supply} and finds that a DQN agent learns a near-optimal policy when other supply chain participants follow a BSP. For a single product, a technique based on semi-Markov average reward to manage inventory decisions across a supply chain is studied in~\cite{giannoccaro2002inventory}. The usage of Q-learning \cite{lillicrap2015continuous} to minimize an operating cost target is proposed in~\cite{kara2018reinforcement} for managing inventory of a single product, and the results are compared against stock-based and age-based policies. Again for a single product on a linear four-echelon supply chain, the authors in~\cite{or-gym} compare different operations research methods with the proximal policy optimization (PPO) algorithm in environments with and without backlogged orders. Their experiments show that PPO outperforms BSP in both environments. An application of the vanilla policy gradient (VPG) algorithm to address a 2-echelon supply chain with stochastic and seasonal demand is discussed in~\cite{peng2019deep}. The quantity of products to ship is determined based on the inventory present in the warehouses. In all experiments, VPG performed better than the $(s, Q)$-policy employed as a baseline.  The authors in~\cite{stranieri2022deep} discuss a single agent RL (SARL) approach to optimize inventory flow across a 2-echelon supply chain. They compare their neuro-dynamic policies to an order-up-to policy, benchmark PPO against the VPG and the A3C algorithms, and conclude that PPO performs the best for managing the inventory. 
 
 The work that is closest to our proposed CMARL and implements a MARL system for IM across a 2-echelon supply chain with 3 stores and 1 warehouse is~\cite{DBLP:journals/corr/abs-2006-04037}. However, there are fundamental differences in the specifications of RL agents with regards to environment dynamics, action spaces, reward structure, and training structure between the two methods. While~\cite{DBLP:journals/corr/abs-2006-04037} assumes the lead times between the warehouse and stores are zero, which is unrealistic as the orders need to be transported and processed, our CMARL explicitly encode the notion of lead time in the environment dynamics. Having a single action policy across all products with all weight parameters shared as in~\cite{DBLP:journals/corr/abs-2006-04037}, results in training time increasing linearly with respect to the number of products. It could reduce performance across multiple products due to catastrophic forgetting~\cite{goodfellow2013empirical}. Our experimental results (refer Fig.~\ref{marl10product}b) indicate that having individual action space for each product has higher reward values alongside the ability to simultaneously train for all products. Unlike~\cite{DBLP:journals/corr/abs-2006-04037} both the warehouse and store agents in CMARL share the system-wide cooperative reward which we demonstrate to be superior, as it avoids sacrificing supply chain-wide globally optimal agent policies for those where agents could compete with each other for local rewards.
 
 In CMARL we further enhance our warehouse agent with extended observation space also comprising of past store actions, and an extended action space to explicitly learn a sub-allocation policy when it has limited inventory to meet all the store requests. This allows all our agents to be trained simultaneously unlike~\cite{DBLP:journals/corr/abs-2006-04037}, which employs a phased training approach and rests on assumptions of unbounded warehouse capacity so that the requested replenishment quantities for all products are always available in the warehouse, while exclusively training the store agents in the first phase. Once every store has converged to a locally optimal policy, only then is the warehouse set to have a finite shelf capacity, and is trained conditioned on converged store agents. However since store agent behavior that is locally optimal may not be globally optimal across a supply chain, this phased training routine results in the warehouse agent being conditioned on a set of subpar policies. Also in~\cite{DBLP:journals/corr/abs-2006-04037}, while the warehouse agent has access to replenishment requests from store agents at the current time period, it does not have access to past replenishment requests. This lack of information results in the warehouse only being able to make a binary decision on whether or not to replenish a store, and cannot intelligently allocate a constrained amount of inventory to stores which our system can achieve. Our CMARL system is able to successfully emulate a divergent supply chain, going beyond the linear supply chains described in most other operations research literature. 	
	
\section{MARL for Inventory Management}
\label{sec:marl}
A Markov Decision Process (MDP) is defined as a tuple  $(\mathcal{S}, \mathcal{A}, \mathcal{T}, \mathcal{R}, \gamma)$ where $\mathcal{S}$ is the state space, $\mathcal{A}$ is the action space, $\mathcal{T}$  is the set of transition probabilities between states, $\mathcal{R}$ is the set of rewards associated with each state, and $\gamma$ is a discount factor. At each time step, the MDP can be completely described by its state $s \in \mathcal{S}$, which is used by an agent to select an action $a \in \mathcal{A}$. According to the set of transition probabilities, the MDP will reach a new state $s^{\prime} \in \mathcal{S}$ in the next time period: $\mathcal{T}(s^{\prime} \in \mathcal{S} | s \in \mathcal{S}, a \in \mathcal{A}): \mathcal{S} \times \mathcal{A} \rightarrow \mathcal{S}$. A RL agent learns a stochastic policy $\pi$ that prescribes the probability of each action $a$ that can be taken in state $s$, as $\pi (a|s) : \mathcal{S} \times \mathcal{A} \rightarrow [ 0, 1 ]$. 

We define a generalization of MDP to a multi-agent setting with the tuple $( \mathcal{U}, \mathbf{S}, \mathbf{A}, \mathbf{T}, \mathbf{R}, \gamma)$. Here, $\mathcal{U} =  \{ u | 1 \leq u \leq |\mathcal{U}| \}$ is the set of agents in the environment, $\mathbf{S}$ is the Cartesian product of the state spaces of all agents $u \in \mathcal{U}$: $\mathbf{S} = \mathcal{S}_{1} \times \mathcal{S}_{2} \times \dots \times \mathcal{S}_{|\mathcal{U}|}$, $\mathbf{A}$ is the Cartesian product of the action spaces of all agents $u \in \mathcal{V}$: $\mathbf{A} = \mathcal{A}_{1} \times \mathcal{A}_{2} \times \dots \times \mathcal{A}_{|\mathcal{U}|}$, $\mathbf{T}$ denotes transition probabilities between $\mathbf{S}$ and $\mathbf{A}$: $\mathbf{T} (\mathbf{s'} \in \mathbf{S} | \mathbf{s} \in \mathbf{S}, \mathbf{a} \in \mathbf{A})$, $\mathbf{R}$ is the Cartesian product over each agent's reward function:  $\mathbf{R} = \mathcal{R}_{1} \times \mathcal{R}_{2} \times \dots \times \mathcal{R}_{|\mathcal{U}|}$, and $\gamma$ is a scalar discount factor. At each time step $t$, all agents synchronously take actions $\mathbf{a} \in \mathbf{A}$. The goal of each agent in the environment is to maximize its long-term reward by finding its own optimal policy $\pi_{u} : \mathcal{S}_{u} \rightarrow \mathcal{A}_{u}$~\cite{sutton2018reinforcement}.
% A multi-agent RL system consists of multiple agents in the same environment, with each agent seeking to maximize its own long-term reward by interacting with the environment and each other. In such a setting, agents can either compete with each other, cooperate with each other or both. Since the objectives of all agents are not necessarily aligned, MARL systems can converge to non-stationary equilibrium points that are not conducive to maximizing system-wide reward. Our solution to this lies in our reward formulation to be described in later sections.
	
\subsection{MARL Implementation}
\begin{figure}[htp]
    \centering
    % \hspace*{-0.5in}
    \begin{subfigure}{.45\textwidth}
        \centering
        \includegraphics[width=\textwidth]{supply_chain_graph}
        %\ctikzfig{supply_chain_graph}
        \caption{Example supply chain topology.}
        \label{topology}
    \end{subfigure}%
    \hspace{5mm}
    \begin{subfigure}{0.45\textwidth}
        \centering
        \includegraphics[width=\textwidth]{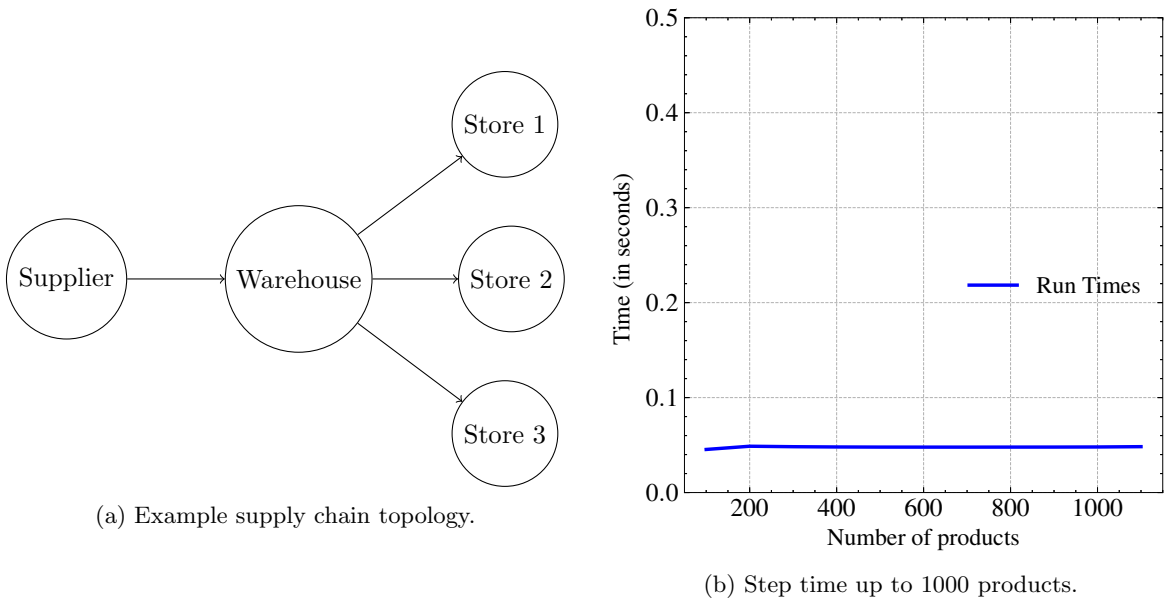}
	\caption{Step time up to 1000 products.}
	\label{step_time}
    \end{subfigure}
    % \vspace*{-3mm}
    \caption{Environment Dynamics.}
\end{figure}

 In the context of our supply chain, $\U = \V \cup \{wh\}$ is the set of vertices with each vertex being either the warehouse agent $(wh)$, or the store agent $v \in \V$ as shown in Fig.~\ref{topology}. While there is a hierarchy present in this supply chain, all agents execute synchronously. Each agent has its own individual policy, state and action spaces, and no agent can directly access the state space of another. \commentbyKG{At each time period, stores request a replenishment quantity from the warehouse agent. Based on this requested quantity and on-hand inventory, the warehouse replenishes each store with an accepted reorder quantity.}
    % \vspace*{-5mm}

% \begin{figure}[h]
%     \centering
%     \ctikzfig{supply_chain_graph}
% \end{figure}
	
\subsection{Store Agents}
Let the vector $\mathbf{x}_v (t) = [x_v(t, 1), \dots, x_v(t, K)]$ denote the on-hand inventory of the $K$ products, and $\mathbf{r}_v(t) = [r_v(t, 1), \dots, r_v(t, K)]$ be the accepted reorder quantity (defined below) at time $t$ for the store agent $v \in \V$. Its state space is defined as: $\mathcal{S}_{v} =  \{\mathbf{x}_{v} (t) \} \cup  \{ \mathbf{r}_v(t - i) \}_{i = 1}^{l_{v}}$, where $l_{v}$ is the lead time to transfer products from the warehouse to store $v$. Since actions taken by an agent $v$ during time period $t$ only affect the reward at time period $t + l_v$ after its corresponding lead time $l_v$, we preserve the Markov property \cite{Miller2001} by accounting for the delay with a history of past replenishment orders included in the observation space (as described in Sec. ~\ref{sec:marl}). Thus, similar to TEXPLORE \cite{hester2013texplore}, agents are able to capture correlations between actions at time $t$ and corresponding rewards at time $t + l_v$.
	
Each store agent learns a policy that performs an action $\mathcal{A}_{v}$ of placing replenishment orders $\mathbf{\hat{r}}_{v} (t) = [\hat{r}_v(t, 1), \dots, \hat{r}_v(t, K)]$ to the warehouse at $t$ for all the $K$ products. We assume that both store and warehouse agents places orders in batches of units, where the action space is quantized into $n$ possible actions for each product, i.e., $\hat{r}_v(t, k) \in \{0, 1, \dots, n \}$. The value $n$ is empirically deduced from historical data.
	
\subsection{Warehouse Agent}
The warehouse agent jointly learns two sub-policies. First, its own replenishment policy for placing replenishment requests $\mathbf{\hat{r}}_{wh}(t)$ from its supplier. Second, an allocation policy for distributing its on-hand inventory to stores $v \in \V$; the accepted reorder quantity $\mathbf{r}_{v} (t)$ which could be less than the requested quantity $\mathbf{\hat{r}}_{v}(t)$, in situations when the warehouse does not have sufficient inventory to meet the requests of all the stores.

The warehouse state space is extended to include replenishment order quantities from stores $\mathbf{\hat{r}}_{v} (t)$, as it enables the warehouse to learn the allocation policy and constrain actual replenishment $\mathbf{\hat{r}}_{v}(t)$ that reaches the stores when needed. Additionally, the state space includes its own inventory $\mathbf{x}_{wh}(t)$ as well as the replenishment orders $\mathbf{r}_{wh}(t)$ placed in the last $m$ time periods to optimally learn a replenishment policy for itself. Its state and actions spaces are:
\begin{align}
    \label{eq:whobservation}
    \mathcal{S}_{wh} &= \{\mathbf{x}_{wh} (t)\} \cup \{ \mathbf{r}_{wh}(t-i) \}^{m}_{i = 1} \cup \{\{\mathbf{\hat{r}}_v (t - i)\}^{m}_{i = 1}| v \in \V\},\\
    \label{eq:whaction}
    \mathcal{A}_{wh} & = \{\mathbf{\hat{r}}_{wh} (t)\} \cup \{ \mathbf{r}_v (t) | v \in \V \}.
\end{align}
In Fig.~\ref{marlsingleproduct}b of Sec.~\ref{sec:experiments}, we observe that when the store replenishment order quantities $\mathbf{\hat{r}}_{v} (t)$ are excluded from the warehouse state space in Eq.~\ref{eq:whobservation}, labelled LimWh-ShRwd, its ability to learn the allocation policy $\mathbf{r}_v (t)$ for each store agent degrades with lower reward value. By additionally observing $\mathbf{\hat{r}}_{v} (t)$, labelled EnWh-ShRwd, the warehouse agent is able to prioritize replenishment to relevant stores. To the best of our knowledge, our work is the first to introduce this specific multi-agent architecture for IM with enhanced state and action spaces for the warehouse agent.

Demand $\mathbf{\hat{r}}_{wh}(t)$ is always met since the supplier is assumed to have infinite capacity, i.e. $ \mathbf{\hat{r}}_{wh}(t) = \mathbf{r}_{wh}(t)$. Ideally, the warehouse should have enough inventory to fulfill all the store requests $\mathbf{\hat{r}}_{v}(t)$. The mechanism for penalizing the system when the warehouse replenishes a smaller amount $\mathbf{r}_{v}(t) \leq \mathbf{\hat{r}}_{v}(t)$ to the stores is described in Sec.~\ref{sec:reward}.
	
\section{Environment for Inventory Management}
\label{sec:environment}
The environment is implemented as a set of tables that keep track of quantities $\mathbf{x}(t), \mathbf{s}(t), \mathbf{\hat{s}}(t), \mathbf{r}(t), \mathbf{\hat{r}}(t)$ and internal tables to keep a log of replenishment order status. These tables are implemented with CuPy \cite{cupy_learningsys2017} which leverages CUDA Toolkit libraries~\cite{cuda} such as cuBLAS, cuRAND, and cuSOLVER to execute matrix operations on a GPU. This allows updates to the environment tables to be execute in constant time (subject to GPU constraints) as shown in Fig.~\ref{step_time}. A detailed description of the environment implementation is provided in Appendix ~\ref{sec:envimplementation}. While the environment can handle an arbitrary number of products, our MARL algorithm implementation converges during training for up to $K=10$ products as discussed in Sec.~\ref{sec:experiments}. We strongly believe the same environment can still be used for our future work when we seek to implement MARL for managing inventory of thousands of products.

A single episode in our environment starts with initial inventory levels at warehouse and store vertices for each product, and executes for $T = 30$ time periods. Each time period may be analogous to a day of sales, where each agent places their respective replenishment order which are received after an associated lead time. Also at each time period, a certain portion of the inventory is sold at the stores according to the customer demand sampled from a demand distribution corresponding to each product.
	
\subsection{Dynamics at Stores}
\label{sec:storedynamics}
 At each time $t$, customer demand $\mathbf{\hat{s}}_{v}(t) = [\hat{s}_v(t, 1) , \dots , \hat{s}_v(t, K)] $ for all the $K$ products is sampled from a product-specific demand distribution. Bounded on $\mathbf{x}_{v}(t)$ the amount of actual inventory sold is: $\mathbf{s}_v(t) = \min\left(\mathbf{\hat{s}}_{v}(t), \mathbf{x}_{v}(t) \right)$, where the $\min()$ function operates element-wise on each product.

At the end of each time period when the store vertex $v$ receives accepted replenishment orders $\mathbf{r}_v(t-l_v)$ placed $l_v$ time periods ago, its inventory gets updated as: $\mathbf{x}_{v}(t + 1) = \mathbf{x}_{v}(t) - \mathbf{s}_{v}(t) + \mathbf{r}_v(t - \mathbf{l}_v)$. Based on $\mathbf{x}_{v}(t)$ and replenishment history $\{ \mathbf{r}_v(t - i) \}_{i = 1}^{l_{v}}$, store $v$ places a replenishment order $\mathbf{\hat{r}}_v(t)$ to the warehouse. We do not need to explicitly model the future demand, as the agent implicitly predicts it based on its learned policy and past demand. We demonstrate this property in Fig.~\ref{marlsingleproduct}b of Sec.~\ref{sec:experiments}, where an \emph{oracle} implementation that can see the true customer demand, $\mathbf{\hat{s}}_{v}\left(t+l_v\right)$, $l_v$ lead time period ahead does not outperform CMARL. The environment enforces that the inventory $\mathbf{x}_{v}(t)$ is always less than the  maximum shelf capacity $\mathbf{c}_v$ of the corresponding product at all times, by setting $\mathbf{x}_{v}(t) = \min\left(\mathbf{x}_{v}(t), \mathbf{c}_v\right)$ element-wise. Inventory that cannot be held in the shelves gets discarded and not realised as sales. However, they need to be procured and stored at the warehouse to be allocated to the stores. Any discarded inventory proportionately penalises the shared reward, defined subsequently in Eq.~\eqref{eq:sharedreward}, by incurring procurement and inventory holding costs at the warehouse without generating any sales revenue. Hence the stores implicitly avoid placing replenishment requests that would result in exceeding their shelf capacity.

% To keep track of replenishment orders placed throughout an episode, the environment keeps track of an 'in-transit' inventory denoted by $\mathbf{x}^{\text{transit}}_{v}(t) $ for each time period $t$. This is the sum of replenishment orders that have been placed and are yet to be delivered. For store at vertex $v\in \mathcal{V} \setminus \{ v_{\text{wh}}\}$, in-transit inventory $\mathbf{x}^{\text{transit}}_{v}(t) = [x^{\text{transit}}_{v}(t, 1) , \dots,  x^{\text{transit}}_{v}(t, K)]$ is updated at the end of each time period as: 

% \begin{equation}
% 	\mathbf{x}^{\text{transit}}_{v}(t + 1) = \mathbf{x}^{\text{transit}}_{v}(t) -  \mathbf{r}_v(t - \mathbf{l}_v) + \mathbf{r}_v(t)
% \end{equation}

% Optionally, the environment can also keep track of unfulfilled orders, defined as the the difference between the demand at a store and the sales quantity at the store. For store at vertex $v \in \mathcal{V} \setminus \{ v_{\text{wh}}\}$, unfulfilled orders $\mathbf{u}_v(t) = [u_v(t, 1) , \dots , u_v(t, K)]$ are calculated as: 

% \begin{equation}
% 	\mathbf{u}_v(t) = \mathbf{\hat{s}}_v(t) - \mathbf{s}_v(t)
% \end{equation}

\subsection{Dynamics at Warehouse}
Based on the store replenishment requests $\mathbf{\hat{r}}_v(t)$, the warehouse uses its allocation sub-policy to decide on the accepted replenishment order $\mathbf{r}_v(t)$. At the end of time $t$, when the warehouse receives its own replenishment $\mathbf{r}_{wh}\left(t-l_{wh}\right)$ placed $l_{wh}$ lead time periods ago, its inventory gets updated as:
\begin{equation*}
    \mathbf{x}_{wh}(t + 1) = \mathbf{x}_{wh}(t) - \left(\sum_{v \in \V} \mathbf{r}_v(t) \right) + \mathbf{r}_{wh}\left(t - l_{wh}\right).
\end{equation*}
Based on its available inventory, the warehouse places a replenishment order $\mathbf{\hat{r}}_{wh}(t)$ to its supplier which is always accepted, as the supplier has no inventory constraint. The maximum shelf capacity is again enforced by the environment by setting $\mathbf{x}_{wh}(t) = \min(\mathbf{x}_{wh}(t), \mathbf{c}_{wh})$ element-wise, where the vector $\mathbf{c}_{wh}$ represents the shelf capacities of each of the products at the warehouse. Any inventory discarded due to shelf capacity constraints, though procured by the warehouse when placing its replenishment request $\mathbf{\hat{r}}_{wh}\left(t - l_{wh}\right)$, are not utilized in allocating to the stores $\mathbf{r}_v(t)$. The system-wide reward defined in Eq.~\eqref{eq:sharedreward} gets penalised proportionally because the cost of procuring this additional inventory is not realized as stores sales. Hence the warehouse agent will implicitly minimize placing surplus replenishment request that cannot be held in its shelves.
%  For warehouse at vertex $v_{\text{wh}}$, in-transit inventory from its supplier $\mathbf{x}^{\text{transit}}_{\text{wh}}(t) = [x^{\text{transit}}_{\text{wh}}(t, 1) , \dots ,  x^{\text{transit}}_{\text{wh}}(t, K)]$ is updated at the end of each time period as: 

% \begin{equation}
% 	\mathbf{x}^{\text{transit}}_{\text{wh}}(t + 1) = \mathbf{x}^{\text{transit}}_{\text{wh}}(t) -  \mathbf{r}_{\text{wh}}(t - \mathbf{l}_{\text{wh}}) + \mathbf{r}_{\text{wh}}(t)
% \end{equation}

% Optionally, the environment can also keep track of unfulfilled replenishment orders, defined as the difference between the amount of replenishment requested by store vertices and the accepted replenishment quantity for  store vertices. For the warehouse vertex $v_{\text{wh}}$, unfulfilled orders $\mathbf{u}_{\text{wh}}(t) = [u_{\text{wh}}(t, 1) , \dots , u_{\text{wh}}(t, K)]$ are calculated as: 

% \begin{equation}
% 	\mathbf{u}_{\text{wh}}(t) = \sum_{v \in \mathcal{V} \setminus \{ v_{\text{wh}}\}} \mathbf{\hat{r}}_{v}(t) - \sum_{v \in \mathcal{V} \setminus \{ v_{\text{wh}}\}} \mathbf{r}_{v}(t)
% \end{equation}

\section{Reward Structure}
\label{sec:reward}
 Since the desired goal is maximizing the system-wide reward (profit across the entire supply chain), situations where agents compete for reward are undesirable. To ensure that the agents are fully cooperative, a shared reward structure is imposed where each agent is rewarded for choosing actions that benefit the system as a whole. This is as opposed to the separate, local reward formulations for store and warehouse agents as specified in~\cite{DBLP:journals/corr/abs-2006-04037}. Sharing rewards can help the learning process converge faster and reach a more optimal solution~\cite{Panait2005CooperativeML}, and reduce the risk of sub-optimal behavior like pursuing individual goals at the cost of the system-wide goal~\cite{omidshafiei2017deep}. Our results in Fig.~\ref{marlsingleproduct}b confirm this advantage of a shared reward structure over individual rewards, and are discussed in Sec.~\ref{sec:experiments}. Our shared reward function consists of the following components:
\subsubsection{Sales Revenue}
As described in Sec.~\ref{sec:storedynamics}, each store sells a certain portion of its inventory $\mathbf{s}_v (t)$ based on customer demand $\mathbf{\hat{s}}_v(t)$. The total revenue made by all the stores is: 
\begin{equation}
\label{eq:sales}
    P_{r} (t) =  \sum^{K}_{k = 1} \sum_{v \in \mathcal{V}} \left[s_v(t, k)  \times \theta_{\text{SP}}(k, v)\right],
\end{equation}
where $\theta_{SP}(k, v)$ is the selling price of a single unit of the $k^{th}$ product at $v$. In most large retailers' supply chains, the the transfer of inventory from the warehouse to stores does not incur an intermediary sale. Thus, we do not explicitly model sales at the warehouse and set its sales revenue to zero. 
	
\subsubsection{Inventory Holding Cost}
Each unit of on-hand inventory at the stores and warehouse typically incur maintenance costs associated with refrigeration, cleaning, electricity etc. Sans this cost, it would be optimal for both warehouse and stores to keep their inventories stocked to near full-capacity at all times by always placing maximal allowed replenishment orders. To discourage such behavior, we introduce a penalty with holding inventory defined as:   
\begin{equation}
\label{eq:holding}
    P_{h}(t) = \sum^{K}_{k = 1}  \sum_{v \in \mathcal{V} \cup {wh}} \left[x_v(t, k) \times \theta_{h}(k, v)\right],
\end{equation}
where $\theta_{h}(k, v)$ is the cost of holding one unit of $k^{th}$ product at $v$.

\subsubsection{Procurement Cost}

As mentioned before, the warehouse procures its replenishment $\mathbf{r}_{wh}(t - l_{wh})$ placed $l_{wh}$ lead time ago from the supplier. The total procurement cost is:
\begin{equation}
\label{eq:procurement}
    P_{p} (t) = \sum^{K}_{k = 1} \left[r_{wh}(t - l_{wh}, k) \times \theta_{CP}(k, wh)\right],
\end{equation}
where $\theta_{\text{CP}}(k, wh)$ is the cost of procuring single unit of product $k$. In most retailers’ supply chains, the inventory transfer from the warehouse to stores does not incur an intermediary sale as they are internal to the organisation. Hence we do not associate procurement cost with the stores.

\subsubsection{Unfulfilled Penalty}

We impose an unfulfilled order penalty when the inventory at the warehouse is insufficient to meet the sum of replenishment requests from stores, and when stores do not have enough inventory to satisfy customer demand. This penalty is formulated as:
\begin{equation}
\label{eq:unfulfilled}
    P_{u}(t) = \theta_{u} \sum_{k=1}^K \text{ReLU} \left(\sum_{v \in \V} \hat{r}_v(t, k) - x_{wh}(t, k) \right) + \theta_u \sum_{k=1}^K \sum_{v \in \V} \left[ \hat{s}_v(t,k) - s_v(t,k) \right],
\end{equation}
where $\theta_{u}$ is a hyper-parameter for this penalty, and $\text{ReLU()}$ is the Rectified Linear Unit function defined as: $\text{ReLU}(x) = \max(0, x)$.

\subsubsection{Shared Reward Specification}
The rewards and penalties in Eqs. ~\ref{eq:sales}-\ref{eq:unfulfilled} are used to calculate a total reward for each time period in an episode, defined as: 
\begin{equation}
\label{eq:sharedreward}
    P(t) = P_{r} (t) - \left(P_{p}(t) + P_{h}(t) + P_{u}(t)\right).
\end{equation}
This expression can also optionally include transportation and labor costs. We exclude them as our experiments deal with a small number of products.
	
\section{Experimental Results}
\label{sec:experiments}
\commentbyKG{
We use a base-stock policy as a baseline to measure the performance of our proposed system. We first demonstrate that a single-agent implementation can match the base-stock policy in rewards on a linear supply chain as described by Hubbs et al.\cite{or-gym}. We then compare a single-agent performance to multi-agent on an identical supply chain. Finally, for a divergent 10-store supply chain, we compare performance between our proposed system and a projected upper bound for the base-stock policy.}

We implement all RL agents with proximal policy optimization (PPO) \cite{schulman2017proximal} over vanilla policy gradient (VPG) as the former is known to have better sample efficiency, improved training stability, more effective exploration of the action space, and robustness to high-dimensional state spaces~\cite{konda1999actor,schulman2015trust,sutton2018reinforcement,silver2014deterministic,lillicrap2015continuous}. A single product is assumed to have a Poisson distribution, with mean parameter $10 \leq \mu \leq 1000$. Each system is trained for $100,000$ episodes on an 8-core vCPU and a single NVIDIA Tesla P4 GPU. We set $\theta_u \gg \theta_{h}$, as this is experimentally determined to minimize stockout occurences while keeping inventory stable\commentbyKG{\footnote{Our non-disclosure agreements precludes us from making the code, deep network architecture, and hyperparameters publicly available.}}.

\subsection{Linear Supply Chain}
We first consider a linear supply chain with 2 vertices $\U = \{wh, v\}$ consisting of one warehouse and one store that deals with a single product. We compare CMARL against inventory control policies such as base-stock policy (BSP)~\cite{anbazhagan2013base} and the single-agent RL (SARL). In the BSP and SARL implementations, the action and observation spaces are global and combined for both the warehouse and the store. The action space is the set of \textit{all} vertices' requested replenishment quantities at time $t$: $\mathcal{A} = \{\mathbf{\hat{r}}_{wh}(t), \mathbf{\hat{r}}_{v}(t)\}$. Likewise, the observation space  is the set of all vertices' on-hand inventories and their past accepted replenishment orders: $\mathcal{S} = \{\mathbf{x}_{wh}(t)\} \cup \{ \mathbf{r}_{wh}(t - i) \}_{i = 1}^{l_{wh}} \} \cup \{\mathbf{x}_{v}(t)\} \cup \{ \mathbf{r}_v(t - i) \}_{i = 1}^{l_{v}}$. The BSP is implemented following the approach described in~\cite{or-gym}.
\commentbyKG{
Following \cite{or-gym}, we implement a derivative free optimization approach using Powell's method \cite{powell1964efficient} to determine a base-stock $\mathbf{z}_w$ quantity for each vertex $w \in \U$. Using $\mathbf{z}_w$ the replenishment requests $\mathbf{\hat{r}}_w(t)$ are calculated as: 
\begin{equation}
    \mathbf{\hat{r}}_w(t) = \max \left(\mathbf{0}, \mathbf{z}_w - \sum_{w \in \mathcal{U}} \left(\mathbf{x}_w(t) + \mathbf{x}^{\text{transit}}_w(t) - \mathbf{u}_w(t) \right)\right)
    \label{eq:basestock}
\end{equation}
where $\mathbf{x}_w(t)$ is the inventory on-hand, $\mathbf{x}^{\text{transit}}_w(t)= \sum\limits_{i=1}^{l_w} \mathbf{r}_{w}(t-i)$ is the in-transit inventory equal to the sum of all accepted replenishment orders placed during the lead time and is yet to arrive, and $\mathbf{u}_w(t)$ denotes the number of orders not fulfilled due to insufficient inventory. Though all the quantities involved are scalar valued corresponding to the single product, we slightly abuse the notation and use boldfaced letters to be consistent.
}

\begin{figure}
    \centering
    \includegraphics[width=1\textwidth]{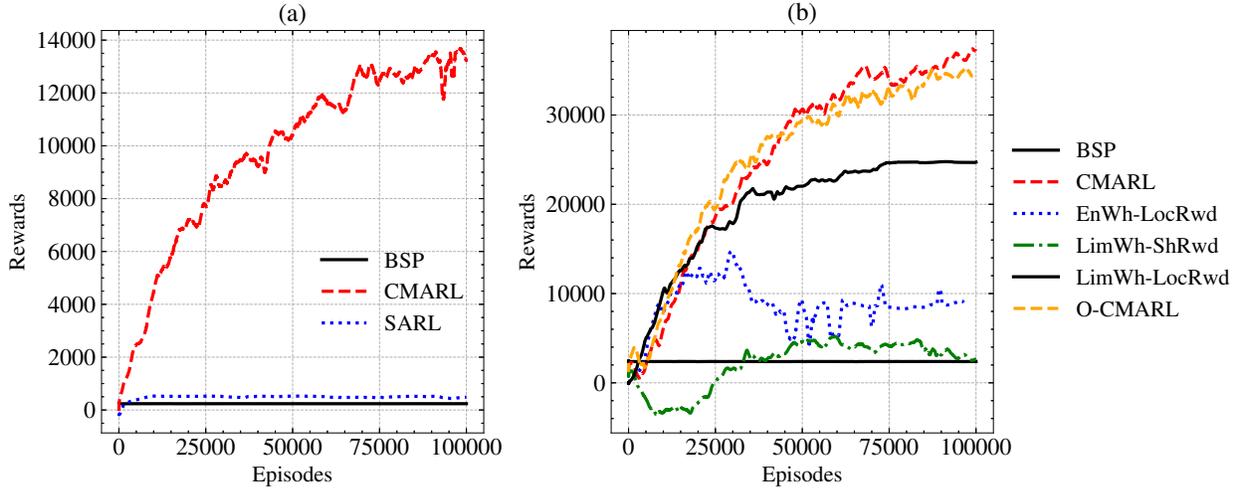}
    % \vspace*{-7mm}

    \caption{Training rewards for: (a) linear supply chain, (b) divergent supply chain with 10 stores, both with 1 product.}
    \label{marlsingleproduct}
\end{figure}

For all the 3 approaches, we compute the rewards using Eq.~\ref{eq:sharedreward} and plot them in Fig.~\ref{marlsingleproduct}a. We observe that the reward from SARL (over 450) is higher compared to the BSP, the latter of which does not exceed 300. This result is supported by \cite{or-gym,DBLP:journals/corr/abs-2006-04037,peng2019deep,oroojlooyjadid2022deep} where SARL, with enough training, is the superior choice to optimization based approaches. However, our proposed CMARL framework far outperforms SARL with more than over 30 times the rewards. The proposed MARL system also reached and surpassed the rewards reaped by the SARL framework in a fraction of the time it took for the latter to reach its maximal value. This is likely because in a multi-agent environment, agents can have different roles and behaviors, leading to a more diverse and efficient exploration of the state space compared to a single agent. As observed in~\cite{lowe2017multi,pmlr-v119-yang20d}, the presence of multiple agents allows for an increase in the effective sample size, leading to faster convergence.\commentbyKG{Also, individual policies have a reduced action and observation space by a factor of $\vert \mathcal{V}\vert$, which decreases the size of the policy function.} 
\commentbyKG{SARL however has drawbacks: the RL agent is still implemented in a centralized manner and so adding or subtracting vertices from the supply chain will require re-training from scratch since there is a singular policy that is being optimized for the whole environment.
}	
% \begin{figure}[h]
%     \centering
%     \begin{subfigure}{.49\textwidth}
%         \centering
%    \includegraphics[width=\textwidth]{Linear_GraphBase.eps}
%     \caption{CMARL rewards vs SARL rewards vs base-stock policy rewards.} 
%     \label{marlsingleproduct}
%     \end{subfigure}%
%     \hfill
%     \begin{subfigure}{0.49\textwidth}
%         \centering
% \includegraphics[width=\textwidth]{10Chains_Graph.eps}
% \caption{Rewards for MARL under different warehouse state and reward specifications. } 
% \label{marlsingleproduct}
%     \end{subfigure}
%     \caption{Training Rewards with a single product.}
%     \label{Training Rewards for 100,000 episodes}
% \end{figure}

\subsection{Divergent Supply Chain}
\subsubsection{Single product}
We now scale the environment to a divergent topology with one warehouse vertex and 10 store vertices, and start with a discussion of our experiments on a single product. An upper bound for the reward for the BSP can be set at $10 \times 360 = 3,600$. This is because a divergent supply chain with $\vert \V \vert$ stores can be modelled as $\vert \V \vert$ linear supply chains with constraints at warehouse vertices. Removing these constraints allows for a larger set of possible behaviors in each of the $\vert \V \vert$ linear supply chains, thereby increasing potential rewards. Over the course of training, average reward reaches over 35,000 which is approximately $10$ times that of the projected reward for base-stock policy, as shown in Fig.~\ref{marlsingleproduct}b. 

We study multiple variants of our MARL system by: (i) limiting the observation space of the warehouse agent defined in Eq.~\eqref{eq:whobservation} to only its past replenishment actions, and excluding the stores' replenishment requests (LimWh), and (ii) having independent localised reward for each agent as opposed to a system-wide shared reward (LocRwd). Specifically, we investigate 5 different MARL configurations: (i) CMARL a.k.a EnWh-ShRwd, (ii) EnWh-LocRwd, (iii) LimWh-ShRwd, (iv) LimWh-LocRwd, and (v) Oracle implementation of CMARL (O-CMARL) where the store agents at time $t$ see the actual customer demand, $\mathbf{\hat{s}}_v(t+l_v)$, $l_v$ lead time periods ahead and make replenishment requests $\mathbf{\hat{r}}_v(t)$ accordingly. As seen in Fig.~\ref{marlsingleproduct}b, our CMARL system that implements both the enhanced observation space for the warehouse agent, and a shared reward structure has the highest reward values. It is not outperformed by the oracle implementation either, implying that we do not explicitly need a forecasting model to foresee the future demand. The agents are able to implicitly predict it based on their learned policy and past demand. The worst performing configuration is the system-wide shared reward without an enhanced warehouse agent (LimWh-ShRwd) which distributes rewards equally, and does not give the system any strong signal on how to improve behavior. As a result allocation decisions are made essentially at random when the warehouse cannot fulfill all replenishment requests and the warehouse fails from learning an efficient allocation policy. Similarly, having an enhanced warehouse with individual rewards (EnWh-LocRwd) for each agent results in the warehouse agent converging to locally advantageous policies much faster than store agents, as it has access to more information about the environment than store agents.

\begin{figure}
    \centering
    \includegraphics[width=1.0\textwidth]{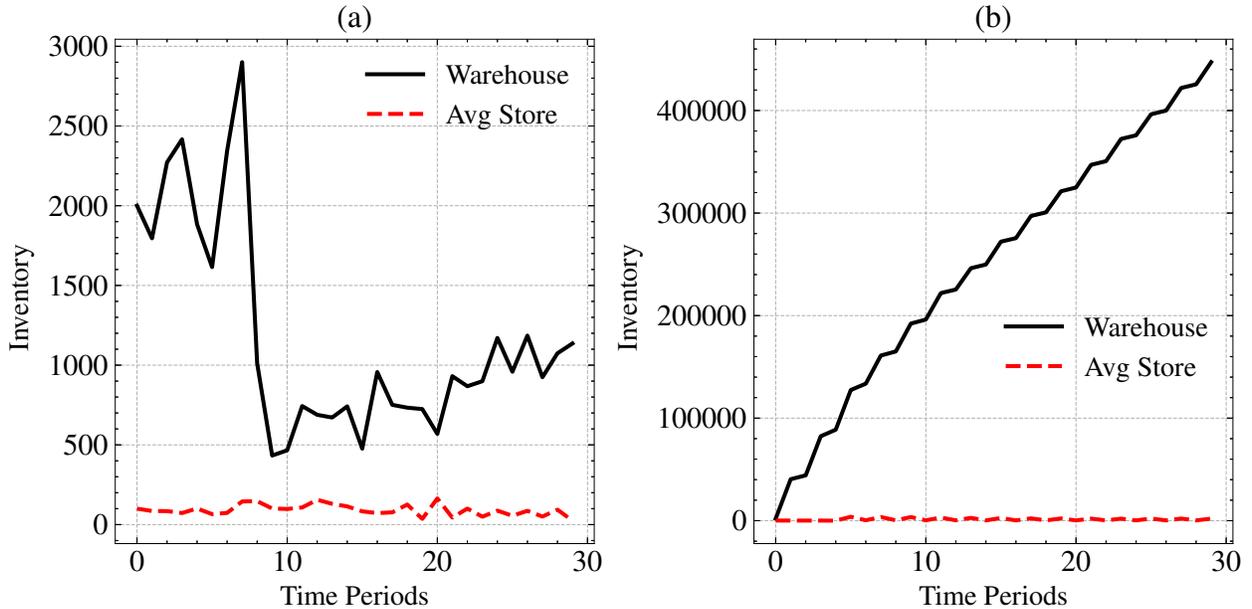}

    % \vspace*{-7mm}
    \caption{Warehouse and average store inventories for one episode (30 time periods) with (a) CMARL, and (b) LimWh-LocRwd for a supply chain with 10 stores, 1 product.}
    \label{inventories}
\end{figure}

A system where the MARL uses neither an enhanced warehouse agent nor a shared reward (LimWh-LocRwd) avoids these issues, as although agents compete for reward they converge to policies more or less at similar rates with respect to each other. Hence it is the second best performer as a system. However, this still produces sub-optimal behavior as depicted in Fig.~\ref{inventories}b, where the inventory levels for each time period are tracked by running the environment for one test episode after the agents policies are converged in the training phase. In the LimWh-LocRwd configuration in Fig.~\ref{inventories}b, the warehouse agent converges to a policy where it simply keeps placing maximal replenishment requests from its supplier, and store agents are erratic with their replenishment requests resulting in frequent stock-outs (as their inventories frequently go to zero). In contrast, the proposed CMARL system manages inventory much better. This is seen in Fig.~\ref{inventories}a, where the warehouse in the CMARL system has a relatively stable inventory that isn't continuously increasing or decreasing. This implies a lowered holding cost as the leftover inventory after sales is minimized, while still fulfilling all orders with inventories never reaching $0$ to have stockouts. 
% \begin{figure}[h!]
% \centering
% \includegraphics[width=\textwidth]{10Chains_Graph.eps}
% \caption{Rewards from our proposed Cooperative MARL framework compared with rewards projected from a base-stock policy for 1 product.} 
% \label{marlsingleproduct}
% \end{figure}

% \begin{figure}[h!]
% \centering
% \includegraphics[width=\textwidth]{Inventory.eps}
% \caption{Warehouse and store inventories over the course of 30 time periods.} 
% \label{inventories}
% \end{figure}
% \begin{figure}[h]
%     \centering
%     \begin{subfigure}{.49\textwidth}
%         \centering
% \includegraphics[width=\textwidth]{Inventory.eps}
% \caption{EnWh-ShRwd (CMARL)} 
% \label{inventories}
%     \end{subfigure}%
%     \hfill
%     \begin{subfigure}{0.49\textwidth}
%         \centering
% \includegraphics[width=\textwidth]{Inventory_False.eps}
% \caption{LimWh-LocRwd} 
% \label{inventories}
%     \end{subfigure}

%     \label{inventories_graphs}
%     \caption{Warehouse and average store inventories for one episode.}
% \end{figure}

     % \vspace*{-7mm}
\subsubsection{Multiple products}
\begin{figure}
    \centering
    \includegraphics[width=1.0\textwidth]{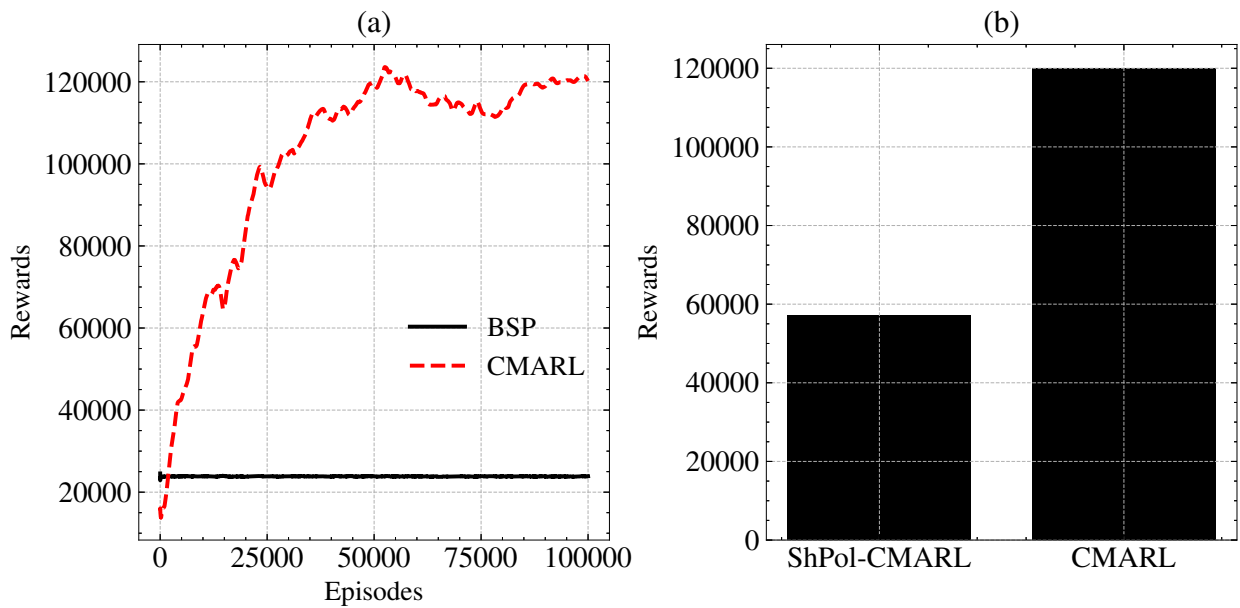}

       % \vspace*{-7mm}
    \caption{(a) Training rewards and (b) Test Reward for one episode for a divergent supply chain with 10 stores, 10 products.}
        \label{marl10product}
\end{figure}

To extend CMARL to supply chains of retailers that deal in multiple products, we experiment with up to 10 products. As described in Sec. \ref{sec:marl}, each agent's action and observation spaces increase linearly with respect to the number of products in the supply chain. This, along with our particular GPU-parallelized environment implementation discussed in Sec. \ref{sec:environment}, enables training the agents policy for multiple products simultaneously thereby avoiding training time overhead, and allows us to independently capture variations in demand distributions for each product individually. Fig.~\ref{marl10product}a shows continuously increasing training rewards for a supply chain that manages inventory of 10 products at each vertex, implying system-wide convergence to optimal policies by agents. 

In contrast, an implementation that shares policy parameters across products such as in \cite{DBLP:journals/corr/abs-2006-04037} needs to be trained separately for each product which increases training time by a factor of $K$. It also results in a relative inability to capture large variations between product types, leading to reduced overall performance. 
As a motivating example, a perishable item such as a fruit has fundamental differences with an electronic device such as a television in trading volume, shelf life, and customer demand so much so that sharing parameters of an inventory management policy for the two is seldom the ideal choice. To emphasize this property, we implemented a variant of CMARL (ShPol-CMARL) with an action and observation space for each agent that can only handle a single product at a time as input, and train it sequentially for each of the 10 products by sharing policy parameters between different products. Fig.~\ref{marl10product}b shows a comparison between rewards for a single episode, between our CMARL system with independent action and state dimensions for each of the $10$ products as described in Secs.~\ref{sec:marl} and \ref{sec:environment}, and the shared policy variant (ShPol-CMARL). We notice CMARL to achieve double the reward of ShPol-CMARL. For the experiments in Fig.~\ref{marl10product}, each one of the Poisson distributions governing product demand had a different value for $\mu$. Our current implementation of CMARL is unable to handle beyond $10$ products for reasons described in the next section.
% \begin{figure}[h]
%     \centering
%     \begin{subfigure}{.49\textwidth}
%         \centering
%     \includegraphics[width=\textwidth]{10SKUs_Graph.eps}
%     \caption{CMARL rewards for 10 products vs projected base-stock policy rewards.} 
%     \label{marl10product}
%     \end{subfigure}%
%     \hfill
%     \begin{subfigure}{0.49\textwidth}
%         \centering
% \includegraphics[width=\textwidth]{marl10product.eps}
% \caption{CMARL trained sequentially for 10 products vs CMARL trained in parallel for 10 products.} 
% \label{marl10product}
%     \end{subfigure}

%     \caption{Rewards for 10 products.}
% \end{figure}

\section{Conclusion and Future Work}
\label{sec:conclusion}
We presented a system with a MARL formulation of the IM problem implemented with a GPU-parallelized environment that consists of one warehouse and multiple stores. Our agent-environment dynamics for the warehouse agent with enhanced observation and action space enables it to effectively learn an allocation sub-policy. Additionally, the shared reward formulation in CMARL encourages cooperation between agents to jointly optimize for a retailer's supply chain needs. We demonstrated that for managing inventory of a single product, CMARL outperforms optimization BSP approaches, single agent RL, as well as other MARL configurations with individual reward structure and limited warehouse observation space.
Our experimental results indicate that having individual action/observation space dimensions corresponding to each product instead of sharing a single policy across all products as done in~\cite{DBLP:journals/corr/abs-2006-04037}, leads to superior reward values alongside the ability to simultaneously train for all products. Our synchronous training of both warehouse and store agents does not require (perhaps unrealistic) assumptions of unlimited warehouse capacity while exclusively training the store agents as performed in phases ~\cite{DBLP:journals/corr/abs-2006-04037}.

For our system, incorporating individual observation and action dimensions for each product in terms of increased reward value, results in the combinatorial explosion of agent's action and observation spaces with increasing numbers of products. This in turn limits the capacity of our current CMARL framework to handle over 10 products, while a IM system for a large retailer requires to handle thousands of products. Our future directions of research include investigating extensions that are equipped to deal with large action spaces such as those proposed in~\cite{farquhar2020growing}, as well as alternatives such as Branching Deep Q-Networks \cite{tavakoli2018action}.

\bibliographystyle{splncs04}
\bibliography{CMARL_biblio}

\appendix
	
\section{Environment Implementation}
\label{sec:envimplementation}

Aside from the primary quantities $\mathbf{x}(t)$,  $\mathbf{s}(t)$, $\mathbf{\hat{s}}(t)$, $\mathbf{r}(t)$, $\mathbf{\hat{r}}(t)$, the environment keeps track of an 'in-transit' inventory denoted by $\mathbf{x}^{\text{transit}}_{v}(t) $ for each time period $t$. This is the sum of replenishment orders that have been placed and are yet to be delivered. For store at vertex $v\in \mathcal{V}$, in-transit inventory $\mathbf{x}^{\text{transit}}_{v}(t) = [x^{\text{transit}}_{v}(t, 1) , \dots,  x^{\text{transit}}_{v}(t, K)]$ is updated at the end of each time period as: 
	
	\begin{equation}
		\mathbf{x}^{\text{transit}}_{v}(t + 1) = \mathbf{x}^{\text{transit}}_{v}(t) -  \mathbf{r}_v(t - l_v) + \mathbf{r}_v(t)
	\end{equation}
	
 For warehouse vertex $wh$, its in-transit inventory from its supplier $\mathbf{x}^{\text{transit}}_{wh}(t)$ is updated at the end of each time period as: 
	
	\begin{equation}
		\mathbf{x}^{\text{transit}}_{wh}(t + 1) = \mathbf{x}^{\text{transit}}_{wh}(t) -  \mathbf{r}_{wh}(t - \mathbf{l}_{wh}) + \mathbf{r}_{wh}(t)
	\end{equation}

 The following CuPy \cite{cupy_learningsys2017} tables (referenced in Sec. \ref{sec:environment}) are constructed to keep track of environment dynamics, where $N = |\mathcal{V}|$ is the number of store vertices: 
	\begin{itemize}
		\item On-hand inventory $I$, in-transit inventory $I_{\text{Transit}}$,  requested reorder Quantities $\hat{R}$ and accepted reorder quantities $R$ with dimensions $(T + 1, N, 2, K)$ such that for $t \in T$:
		\begin{equation}
			I(t) = \begin{bmatrix} 
				\mathbf{x}_{wh}(t) & \mathbf{x}_{v_{1}}(t)  \\
				\vdots & \vdots \\
				\mathbf{x}_{wh}(t) & \mathbf{x}_{v_{N}}(t)    
			\end{bmatrix} \text{ and  }
			I_{\text{Transit}}(t) = \begin{bmatrix} 
				\mathbf{x}^{\text{transit}}_{wh}(t) & \mathbf{x}^{\text{transit}}_{v_{1}}(t)  \\
				\vdots & \vdots \\
				\mathbf{x}^{\text{transit}}_{wh}(t) & \mathbf{x}^{\text{transit}}_{v_{N}}(t)    
			\end{bmatrix}
		\end{equation}

		\begin{equation}
			\hat{R}(t) =\begin{bmatrix} 
				\mathbf{\hat{r}}_{wh}(t) & \mathbf{\hat{r}}_{v_{1}}(t)  \\
				\vdots & \vdots \\
				\mathbf{\hat{r}}_{wh}(t) & \mathbf{\hat{r}}_{v_{N}}(t)    
			\end{bmatrix} \text{ and  }  R(t) = \begin{bmatrix} 
				\mathbf{r}_{wh}(t) & \mathbf{r}_{v_{1}}(t)  \\
				\vdots & \vdots \\
				\mathbf{r}_{wh}(t) & \mathbf{r}_{v_{N}}(t)    
			\end{bmatrix}
		\end{equation}
		
		where for $v \in \mathcal{V}$, $\mathbf{x}_{v}(t) = [x_v(t, 1), \dots,  x_v(t, K)]$, $\mathbf{x}^{\text{transit}}_{v}(t) = [x^{\text{transit}}_v(t, 1), \dots,  x^{\text{transit}}_v(t, K)]$, $\mathbf{\hat{r}}_{v}(t) = [\hat{r}_v(t, 1), \dots, \hat{r}_v(t, K)]$ and $\mathbf{r}_{v}(t) = [r_v(t, 1), \dots, r_v(t, K)]$.
	
    \item Maximum inventory capacity $C$ of shape $(N, 2, K)$ such that:
    \begin{equation}
        C = \begin{bmatrix} 
            \mathbf{c}_{wh} & \mathbf{c}_{v_{1}}  \\
            \vdots & \vdots \\
            \mathbf{c}_{wh} & \mathbf{c}_{v_{N}}    
        \end{bmatrix}
    \end{equation}
    where for $v \in \mathcal{V}$, $\mathbf{c}_{v} = [c_v(1), \dots, c_v(K)]$.

		\item Store vertex demand $D$ of shape $(T, N, K)$ such that for $t \in T$: 
		
		\begin{equation}
			D(t) = \begin{bmatrix} 
				\mathbf{\hat{s}}_{v_{1}}(t)  \\
				\vdots \\
				\mathbf{\hat{s}}_{v_{N}}(t)    
			\end{bmatrix}
		\end{equation}
		
		where for $v \in \mathcal{V}$, $\mathbf{\hat{s}}_{v}(t) = [\hat{s}_v(t, 1), \dots, \hat{s}_v(t, K)]$.
		
		\item Vertex sales $S$ of shape $(T, N, 2, K)$ such that for $t \in T$: 
		\begin{equation}
			S(t) =  \begin{bmatrix} 
				\mathbf{r}_{v_{1}}(t) & \mathbf{s}_{v_{1}}(t) \\
				\vdots & \vdots \\
				\mathbf{r}_{v_{N}}(t)  & \mathbf{s}_{v_{N}}(t)  
			\end{bmatrix} 
		\end{equation}
		
		where for $v \in \mathcal{V}$, $\mathbf{s}_{v}(t) = [s_v(t, 1), \dots, s_v(t, K)]$, $\mathbf{r}_{v}(t) = [r_v(t, 1), \dots, r_v(t, K)]$.

		\item Reward $P \in \mathbb{R}^T$ such that $P(t)$ (described in Section 5 of the main paper) is the reward for the time period $t$.
	   \end{itemize} 
	
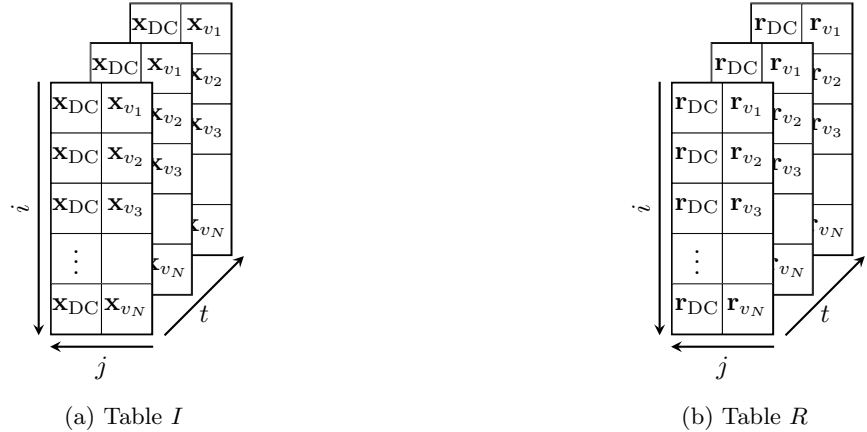
\begin{figure*}[h]

    \begin{center}
    \begin{subfigure}[b]{.5\textwidth}
  \centering

 \begin{tikzpicture}[auto matrix/.style={matrix of nodes,
  draw,thick,inner sep=0pt,
  nodes in empty cells,column sep=-0.2pt,row sep=-0.2pt,
  cells={nodes={minimum width=1.9em,minimum height=1.9em,
   draw,very thin,anchor=center,fill=white,
   execute at begin node={%
   $\vphantom{x_|}\ifnum\the\pgfmatrixcurrentrow<4
     \ifnum\the\pgfmatrixcurrentcolumn<2
      {#1}_{\text{DC}}
     \else 
      \ifnum\the\pgfmatrixcurrentcolumn=2
       {#1}_{v_{\the\pgfmatrixcurrentrow}}
      \fi
     \fi
    \else
     \ifnum\the\pgfmatrixcurrentrow=5
      \ifnum\the\pgfmatrixcurrentcolumn<2
       {#1}_{\text{DC}}
      \else
       \ifnum\the\pgfmatrixcurrentcolumn=2
        {#1}_{v_{N}}
       \fi 
      \fi
     \fi
    \fi  
    \ifnum\the\pgfmatrixcurrentrow\the\pgfmatrixcurrentcolumn=14
     \cdots
    \fi
    \ifnum\the\pgfmatrixcurrentrow\the\pgfmatrixcurrentcolumn=41
     \vdots
    \fi
    \ifnum\the\pgfmatrixcurrentrow\the\pgfmatrixcurrentcolumn=44
     \ddots
    \fi$
    }
  }}}]
 \matrix[auto matrix=\mathbf{x},xshift=3em,yshift=3em](matz){
  &   \\
  &  \\
  &   \\
  &   \\
  &   \\
 };
 \matrix[auto matrix=\mathbf{x},xshift=1.5em,yshift=1.5em](maty){
  &   \\
  &   \\
  &   \\
  &   \\
  &   \\
 };
 \matrix[auto matrix=\mathbf{x}](matx){
  &  \\
  &  \\
  &  \\
  &  \\
  &  \\
 };
 \draw[thick,-stealth] ([xshift=1ex]matx.south east) -- ([xshift=1ex]matz.south east)
  node[midway,below] {$t$};
 \draw[thick,stealth-] ([yshift=-1ex]matx.south west) -- 
  ([yshift=-1ex]matx.south east) node[midway,below] {$j$};
 \draw[thick,-stealth] ([xshift=-1ex]matx.north west)
   -- ([xshift=-1ex]matx.south west) node[midway,above,rotate=90] {$i$};
\end{tikzpicture}

  \caption{Table $I$}
  \label{fig:sub1}
\end{subfigure}%
\begin{subfigure}[b]{.5\textwidth}
  \centering
   \begin{tikzpicture}[auto matrix/.style={matrix of nodes,
  draw,thick,inner sep=0pt,
  nodes in empty cells,column sep=-0.2pt,row sep=-0.2pt,
  cells={nodes={minimum width=1.9em,minimum height=1.9em,
   draw,very thin,anchor=center,fill=white,
   execute at begin node={%
   $\vphantom{x_|}\ifnum\the\pgfmatrixcurrentrow<4
     \ifnum\the\pgfmatrixcurrentcolumn<2
      {#1}_{\text{DC}}
     \else 
      \ifnum\the\pgfmatrixcurrentcolumn=2
       {#1}_{v_{\the\pgfmatrixcurrentrow}}
      \fi
     \fi
    \else
     \ifnum\the\pgfmatrixcurrentrow=5
      \ifnum\the\pgfmatrixcurrentcolumn<2
       {#1}_{\text{DC}}
      \else
       \ifnum\the\pgfmatrixcurrentcolumn=2
        {#1}_{v_{N}}
       \fi 
      \fi
     \fi
    \fi  
    \ifnum\the\pgfmatrixcurrentrow\the\pgfmatrixcurrentcolumn=14
     \cdots
    \fi
    \ifnum\the\pgfmatrixcurrentrow\the\pgfmatrixcurrentcolumn=41
     \vdots
    \fi
    \ifnum\the\pgfmatrixcurrentrow\the\pgfmatrixcurrentcolumn=44
     \ddots
    \fi$
    }
  }}}]
 \matrix[auto matrix=\mathbf{r},xshift=3em,yshift=3em](matz){
  &   \\
  &  \\
  &   \\
  &   \\
  &   \\
 };
 \matrix[auto matrix=\mathbf{r},xshift=1.5em,yshift=1.5em](maty){
  &   \\
  &   \\
  &   \\
  &   \\
  &   \\
 };
 \matrix[auto matrix=\mathbf{r}](matx){
  &  \\
  &  \\
  &  \\
  &  \\
  &  \\
 };
 \draw[thick,-stealth] ([xshift=1ex]matx.south east) -- ([xshift=1ex]matz.south east)
  node[midway,below] {$t$};
 \draw[thick,stealth-] ([yshift=-1ex]matx.south west) -- 
  ([yshift=-1ex]matx.south east) node[midway,below] {$j$};
 \draw[thick,-stealth] ([xshift=-1ex]matx.north west)
   -- ([xshift=-1ex]matx.south west) node[midway,above,rotate=90] {$i$};
\end{tikzpicture}
  \caption{Table $R$}
  \label{fig:sub2}
\end{subfigure}

\end{center}
    \caption{Illustration of environment table structure with 2 examples.}
    \label{fig:my_label}
\end{figure*}

	 With reference to the topology described in Fig 1(a) of the main paper, each graph walk from the warehouse to a store is a row in the aforementioned tables (except for $P$). In each row, the corresponding value for the warehouse vertex $wh$ is copied $N$ times such that the entire environment executes as a set of $N$ linear supply chains.  Fig.~\ref{fig:sub1} and \ref{fig:sub2} illustrate the environment's table structure with the entirety of tables $I$ and $R$ for one episode.

\end{document}